\UseRawInputEncoding
\documentclass{llncs}
\usepackage{llncsdoc}
\usepackage{epsfig}
\usepackage{booktabs}
\usepackage{wrapfig}
\usepackage{graphicx}
\usepackage{cite}
\usepackage{booktabs}
\usepackage{array}
\usepackage{algorithm}
\usepackage{algorithmic}
\usepackage{graphicx}
\usepackage{amsmath}
\usepackage{amssymb}
\usepackage{relsize}
\usepackage{subcaption}
\usepackage{fancyhdr}
\usepackage{color}

\usepackage[retainorgcmds]{IEEEtrantools}
\DeclareMathOperator*{\argmax}{arg\,max}
\DeclareMathOperator*{\argmin}{arg\,min}
\DeclareMathOperator*{\tr}{Tr}
\newtheorem{prop}{Proposition}
\newtheorem{thm}{Theorem}

\begin{document}
\title{Direct Estimation of Spinal Cobb Angles by Structured Multi-Output Regression}
\author{Haoliang Sun$^{1,2}$, Xiantong Zhen$^{2}$, Chris Bailey$^{3}$, Parham Rasoulinejad$^{3}$, Yilong Yin$^{1}$, and Shuo Li$^{2}$}

\institute{$^{1}$ Shandong University, Jinan, China\\
$^{2}$ The University of Western Ontario, London, ON, Canada\\
$^{3}$ London Health Sciences Center, ON, Canada}

\maketitle

\thispagestyle{fancy}            
\fancyhead{}                     
\rhead{}    
\chead{\bf\textit{\color{blue}{Proceedings of International Conference on Information Processing in Medical Imaging (IPMI 2017)}}}              
\lfoot{}                 
\cfoot{}                    

\renewcommand{\headrulewidth}{0pt}      
\renewcommand{\footrulewidth}{0pt}

\pagestyle{empty}

\begin{abstract}
The Cobb angle that quantitatively evaluates the spinal curvature plays an important role in the scoliosis diagnosis and treatment. Conventional measurement of these angles suffers from huge variability and low reliability due to intensive manual intervention. However, since there exist high ambiguity and variability around boundaries of vertebrae, it is challenging to obtain Cobb angles automatically. In this paper, we formulate the estimation of the Cobb angles from spinal X-rays as a multi-output regression task. We propose structured support vector regression (S$^2$VR) to jointly estimate Cobb angles and landmarks of the spine in X-rays in one single framework. The proposed S$^2$VR can faithfully handle the nonlinear relationship between input images and quantitative outputs, while explicitly capturing the intrinsic correlation of outputs. We introduce the manifold regularization to exploit the geometry of the output space. We propose learning the kernel in S$^2$VR by kernel target alignment to enhance its discriminative ability. The proposed method is evaluated on the spinal X-rays dataset of 439 scoliosis subjects, which achieves the inspiring correlation coefficient of $92.76\%$ with ground truth obtained manually by human experts and outperforms two baseline methods. Our method achieves the direct estimation of Cobb angles with high accuracy, which indicates its great potential in clinical use.
\end{abstract}

\section{Introduction}
\vspace{-3mm}

\begin{figure}[!htb]
    \centering
    \begin{tabular}[t]{cc}

\begin{subfigure}{0.4\textwidth}
    \centering
    \smallskip
    \includegraphics[width=2.167cm,height=5.5cm]{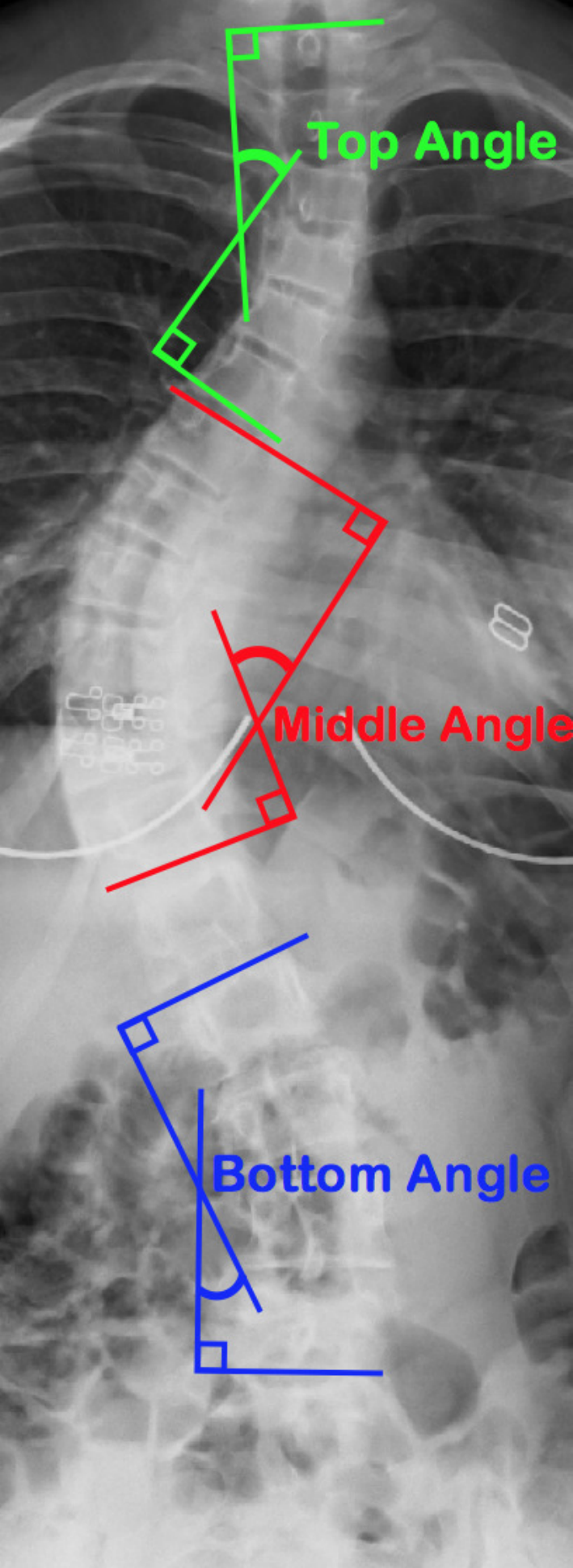}
    \caption{}
    \label{cobb}
\end{subfigure}
    &
        \begin{tabular}{cccc}
        \smallskip
            \begin{subfigure}[t]{0.1\textwidth}
                \centering
                \includegraphics[width=0.9362cm,height=2.35cm]{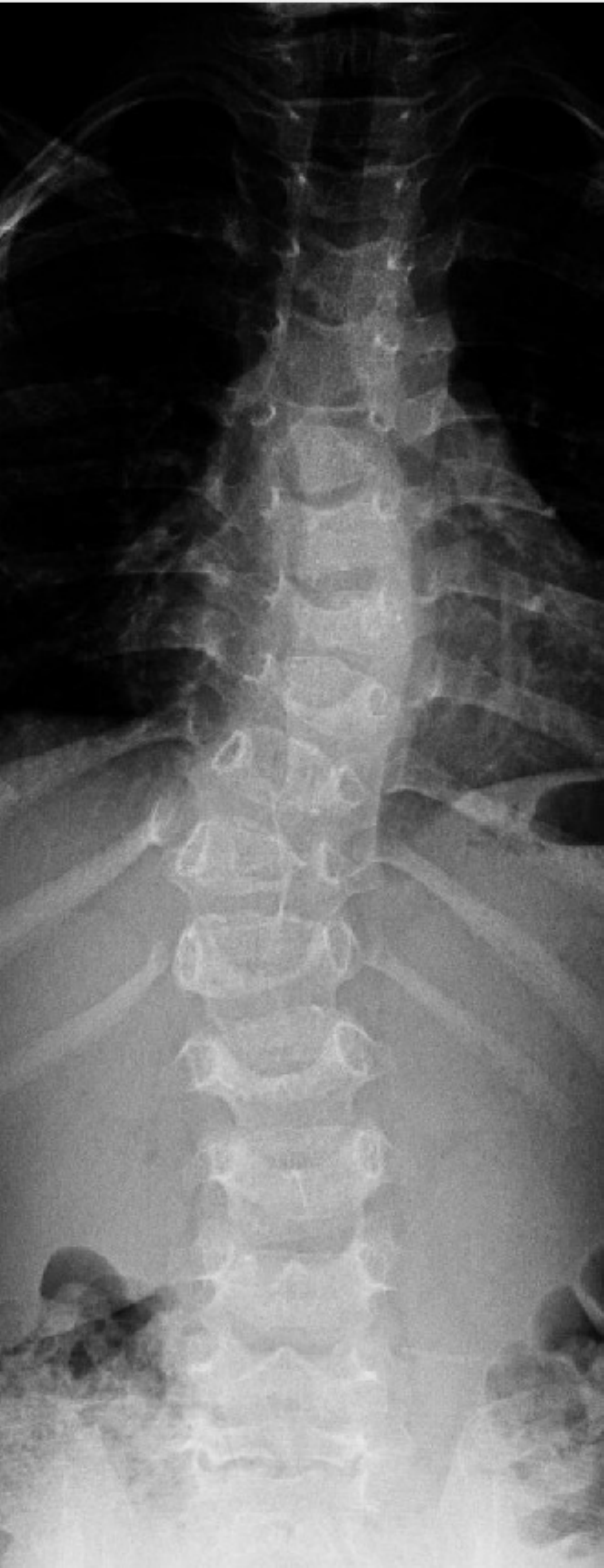}
                \caption{}
            \end{subfigure}&

            \begin{subfigure}[t]{0.1\textwidth}
                \centering
                \includegraphics[width=0.9362cm,height=2.35cm]{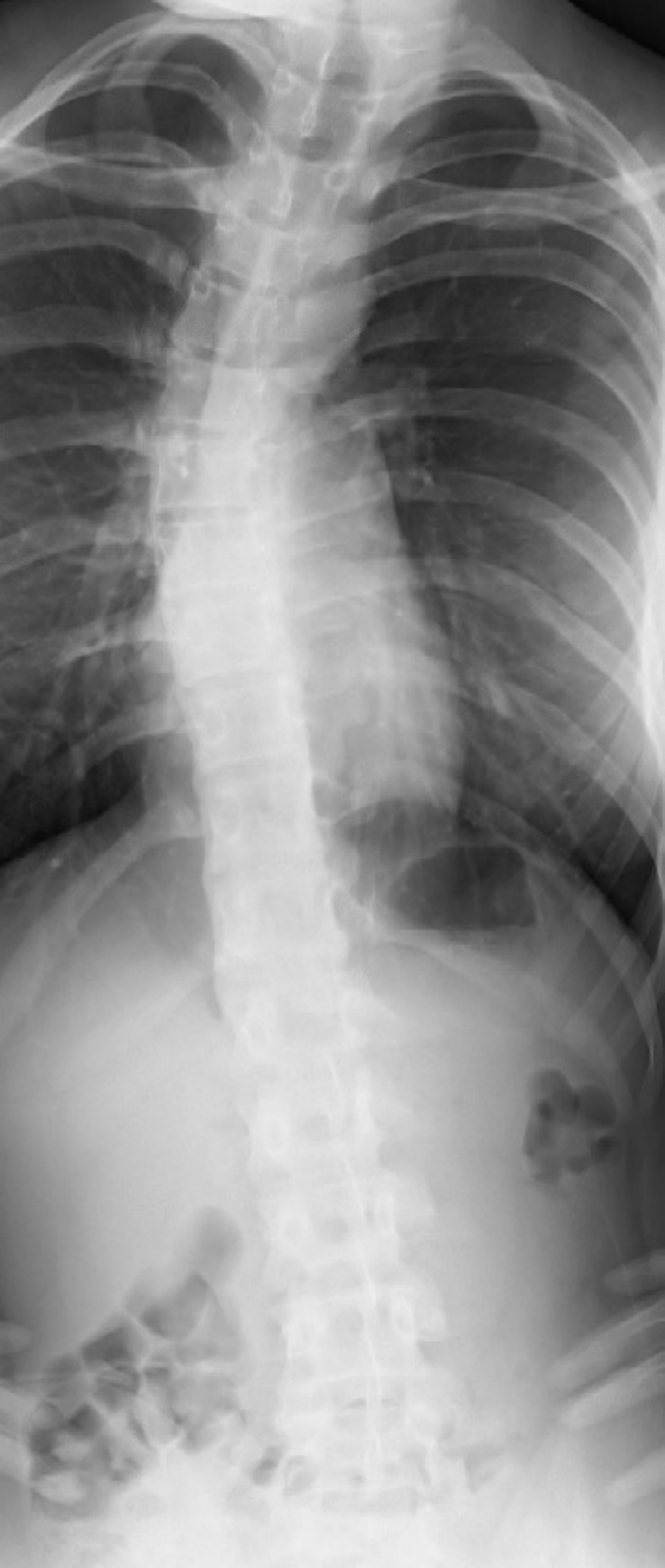}
                \caption{}
            \end{subfigure}&

            \begin{subfigure}[t]{0.1\textwidth}
                \centering
                \includegraphics[width=0.9362cm,height=2.35cm]{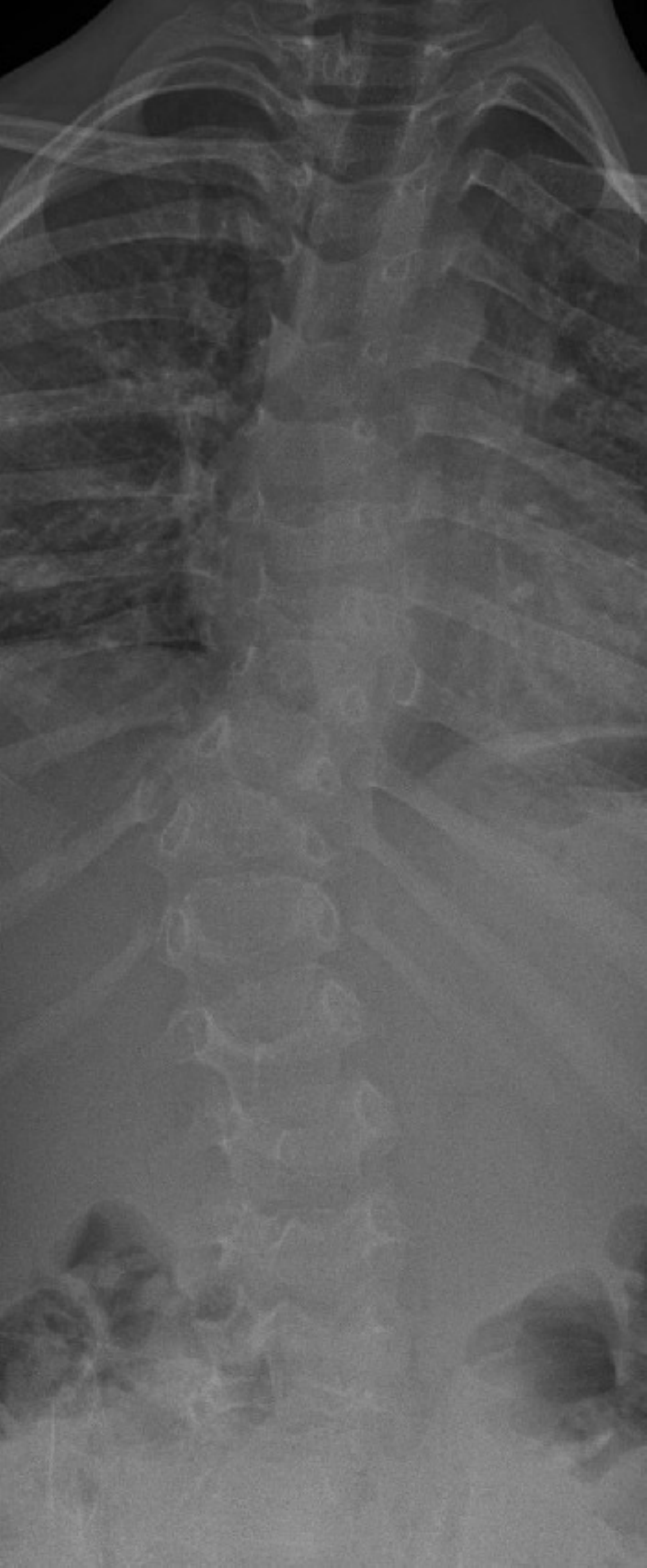}
                \caption{}
            \end{subfigure}&

            \begin{subfigure}[t]{0.1\textwidth}
                \centering
                \includegraphics[width=0.9362cm,height=2.35cm]{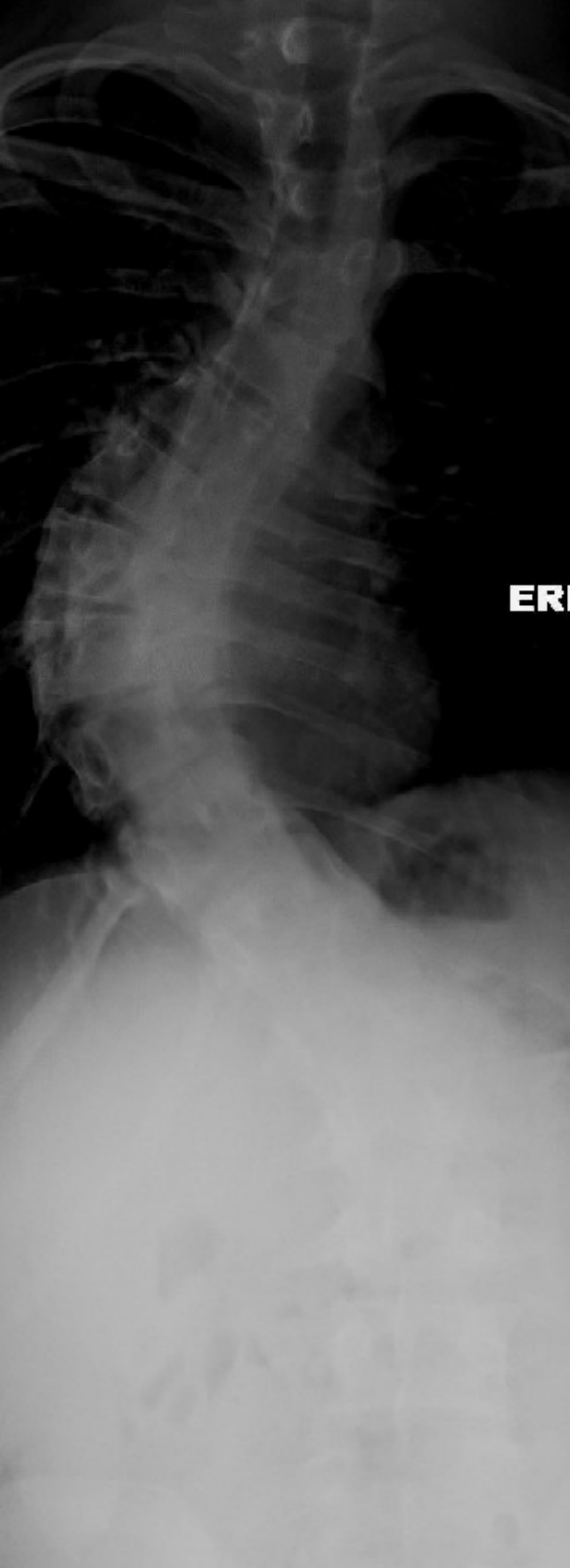}
                \caption{}
            \end{subfigure}\\

            \begin{subfigure}[t]{0.1\textwidth}
                \centering
                \includegraphics[width=0.9062cm,height=2.3cm]{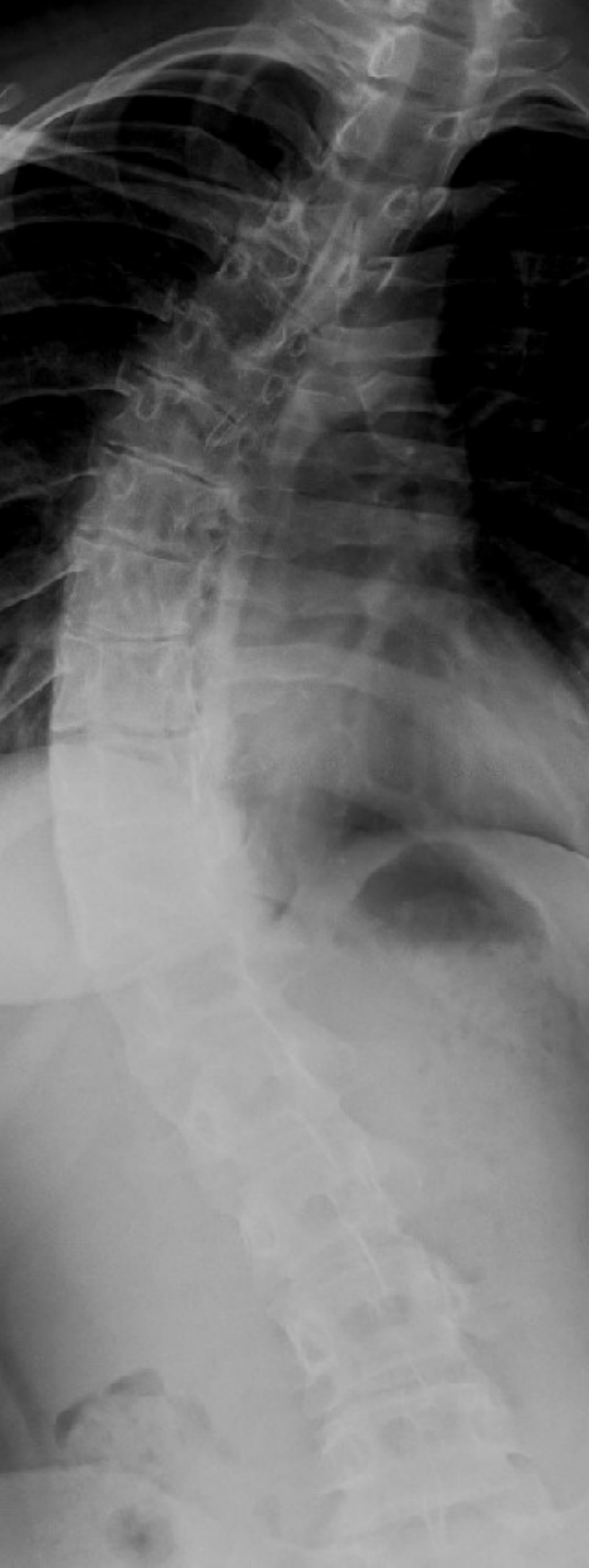}
                \caption{}
            \end{subfigure}&

            \begin{subfigure}[t]{0.1\textwidth}
                \centering
                \includegraphics[width=0.9062cm,height=2.3cm]{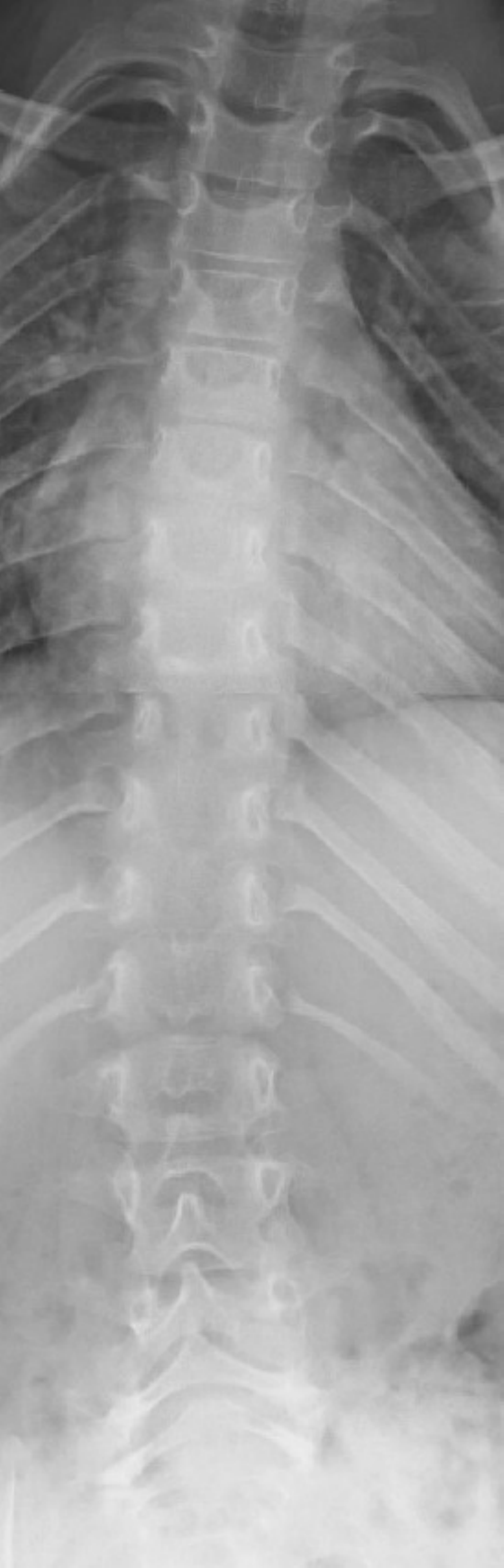}
                \caption{}
            \end{subfigure}&

            \begin{subfigure}[t]{0.1\textwidth}
                \centering
                \includegraphics[width=0.9062cm,height=2.3cm]{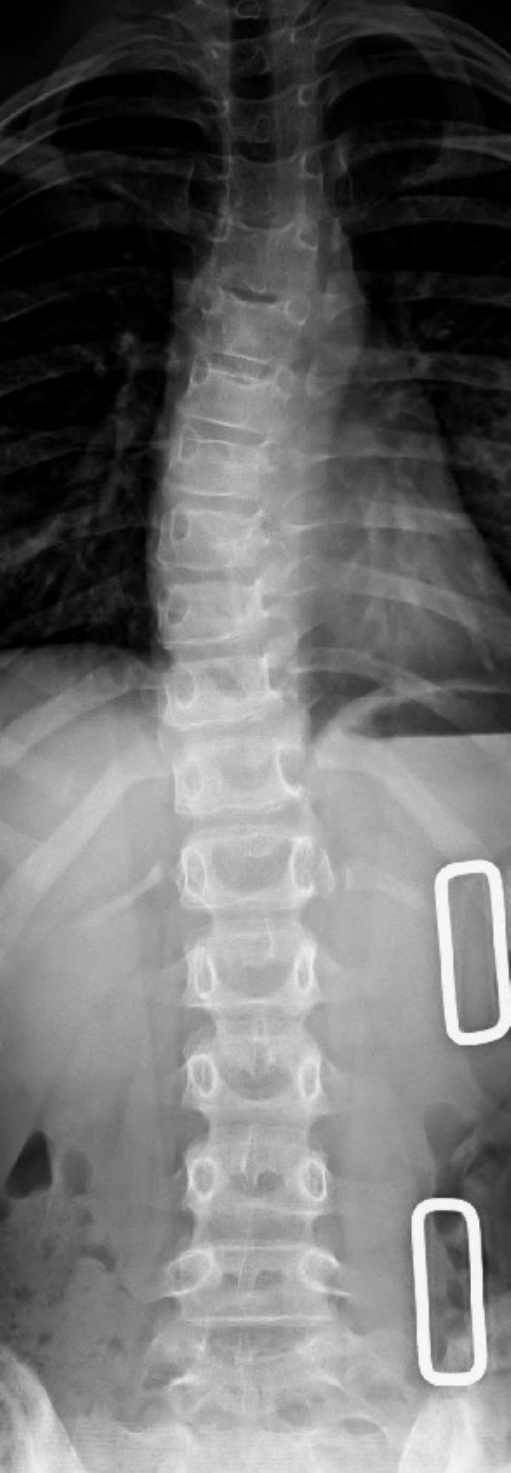}
                \caption{}
            \end{subfigure}&

            \begin{subfigure}[t]{0.1\textwidth}
                \centering
                \includegraphics[width=0.9062cm,height=2.3cm]{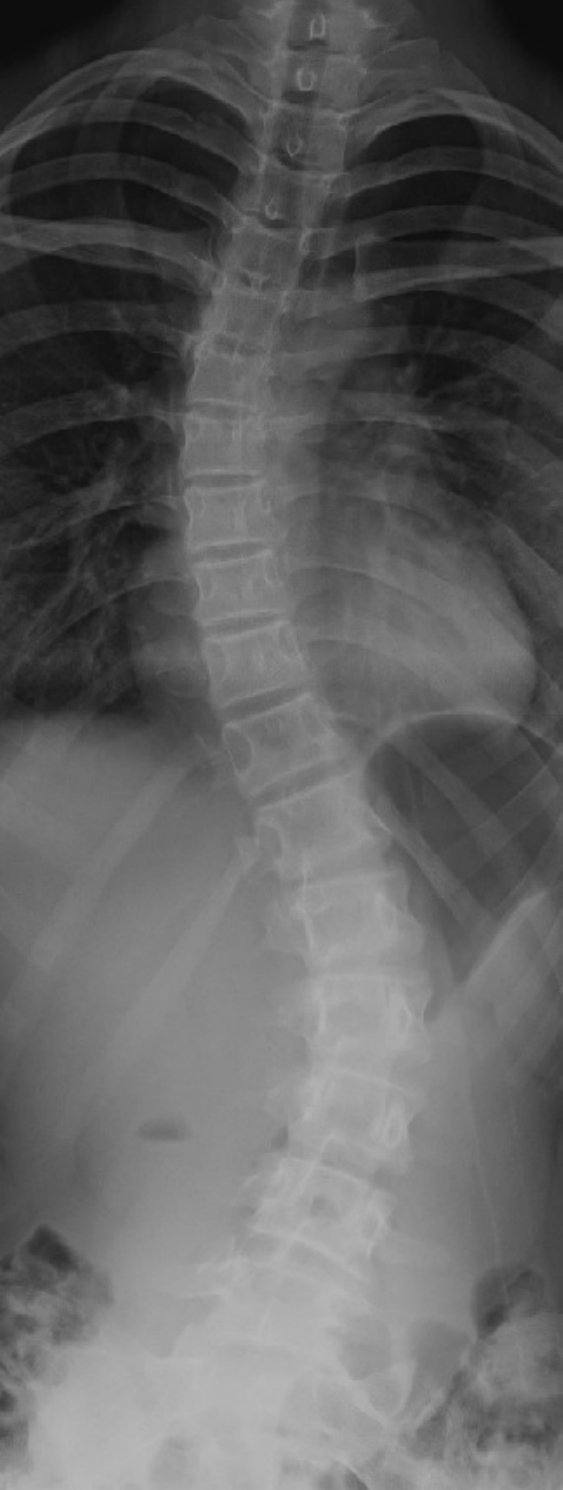}
                \caption{}
            \end{subfigure}
        \end{tabular}

    \end{tabular}
    \vspace{-3mm}
    \caption{It is challenging to measure three Cobb angles in (a) due to high ambiguity and variability in scoliosis X-rays from different subjects (a-g).}
    \label{demo}
    \vspace{-6mm}
\end{figure}

Cobb angles, which are manually measured on X-rays, are widely used for scoliosis diagnosis and treatment decisions. Scoliosis is a structural, lateral, rotated curvature of the spine, which especially arises in children at or around puberty and leads to disability \cite{weinstein2008adolescent}. For clinical examination of scoliosis, the radiography (X-ray) is the most common imaging technique with the cheap acquisition and less time cost \cite{greiner2002adolescent}. Cobb angles (Fig~\ref{cobb}) derived from a posteroanterior (back to front) X-ray and measured by selecting the most tilted vertebra at the top and bottom of the spine respect to the horizontal line are faithfully used to quantify the magnitude of spinal deformities \cite{vrtovec2009review}.

However, conventional manual measurement involves the heavy intervention of identifying the vertebrae and measuring angles, it suffers from huge variability and high unreliability while being labor-intensive. The accuracy of measurement is affected by many factors, such as the selection of vertebrae, the bias of different observers and inaccurate protractors.

It is challenging to estimate Cobb angles automatically due to the high ambiguity and variability of X-rays. As shown in Fig~\ref{demo}, large anatomical variability and low tissue contrast can lead to the complexity of identification of interesting vertebrae and further measurement. Active contour model \cite{anitha2012automatic}, customized filter \cite{anitha2014automatic} and charged-particle model \cite{sardjono2013automatic} were used for localizing vertebrae. Angles are computed from the slope of these vertebrae. Although these methods can derive the Cobb angles automatically, they suffer from the complexity of multiple processing stages, which is computationally expensive. Moreover, segmentation for multiple objects of all vertebrae is extremely challenging, especially in spinal X-rays with variant structures and ambiguous boundaries.

Without filtering or segmentation, direct estimation methods \cite{afshin2012global, wang2014direct,zhen2015direct,zhen2016multi,zhen2017direct} based on statistical learning, which achieves the great success of large-scale image data recently, have the potential for Cobb angles estimation. Compared with conventional segmentation methods, such as variational models \cite{mumford1989optimal}, direct estimation methods characterize the relationship between the appearance of images and high-level semantic concepts via supervised learning, which enables efficient computation and compact structure modeling. By exploring data statistics, direct methods are robust and can handle fuzzy boundaries and region heterogeneity.

The estimation of the Cobb angles can benefit from the prediction of coordinates of landmarks. Due to the ambiguity and variability of spinal images, it is challenging to obtain the angles separately. Intuitively, the Cobb angles closely correspond to the spinal shape which can be substantially characterized by landmarks. In this paper, we train the model by minimizing the loss with respect to both the Cobb angles and all coordinates of landmarks simultaneously, which demonstrates the robustness of our method. In contrast to existing direct methods \cite{afshin2012global, wang2014direct,zhen2015direct,zhen2016multi}, the proposed method jointly estimates landmarks and Cobb angles, which not only improves the prediction accuracy but also provides an intuitive validation of angle estimation for the radiologist.

Although existing direct methods achieved great success, their direct application to our task suffers from three major limitations. First, these methods fail to explicitly characterize the correlations among outputs, which is essential for discriminative learning. Second, they neglect the intrinsic geometry of the distribution of output values, which would result in the non-smooth solution and then has a negative effect on the angle estimation. Third, the kernel parameters, e.g., the bandwidth, are usually set manually, which does not guarantee to be optimal for different applications.

In this paper, we propose structured support vector regression (S$^2$VR) to directly predict the Cobb angles and landmarks of a spine from its X-ray. In comparison to conventional support vector regression (SVR), the S$^2$VR accomplishes of nonlinear mapping and explicit correlation modeling in one single framework. In particular, the nonlinear mapping handles the relationship between input images and high-level outputs (angles and landmarks), while the explicit correlation modeling captures the correlations among outputs. 

In summary, our work contributes to the following aspects:
\begin{itemize}
    \item We achieve the jointly automatic estimation of Cobb angles and spinal landmarks by multi-output regression, which is essential for the evaluation of spinal curvature from radiography directly.
    \item We propose a novel multi-output regression model called S$^2$VR, which can simultaneously handle the nonlinear input-output relationship and intrinsic inter-output correlation. The framework is endowed with great generality which enables wide applications for other measurement estimation.
    \item We introduce a manifold regularizer into the framework by exploring the geometry structure of the output space, which enables smooth solutions.
    \item To improve the performance, we propose to learn the kernels by using kernel target alignment, which increase the discriminative ability of the kernels.
\end{itemize}

\vspace{-5mm}
\section{Methodology}

The proposed structured support vector regression (S$^2$VR) directly estimates the Cobb angles and coordinates of landmarks by formulating the prediction as a multi-output regression task. S$^2$VR consists of nonlinear mapping process and explicit correlation learning stage. The manifold regularization is introduced to explore the intrinsic geometry of the output space. Since our method is built on support vector regression, we learn the kernel based on kernel target alignment.

\subsection{Structured Support Vector Regression}
\begin{figure}[t]
   \vspace{-7mm}
\begin{center}
   \centerline{\includegraphics[width=0.8\linewidth]{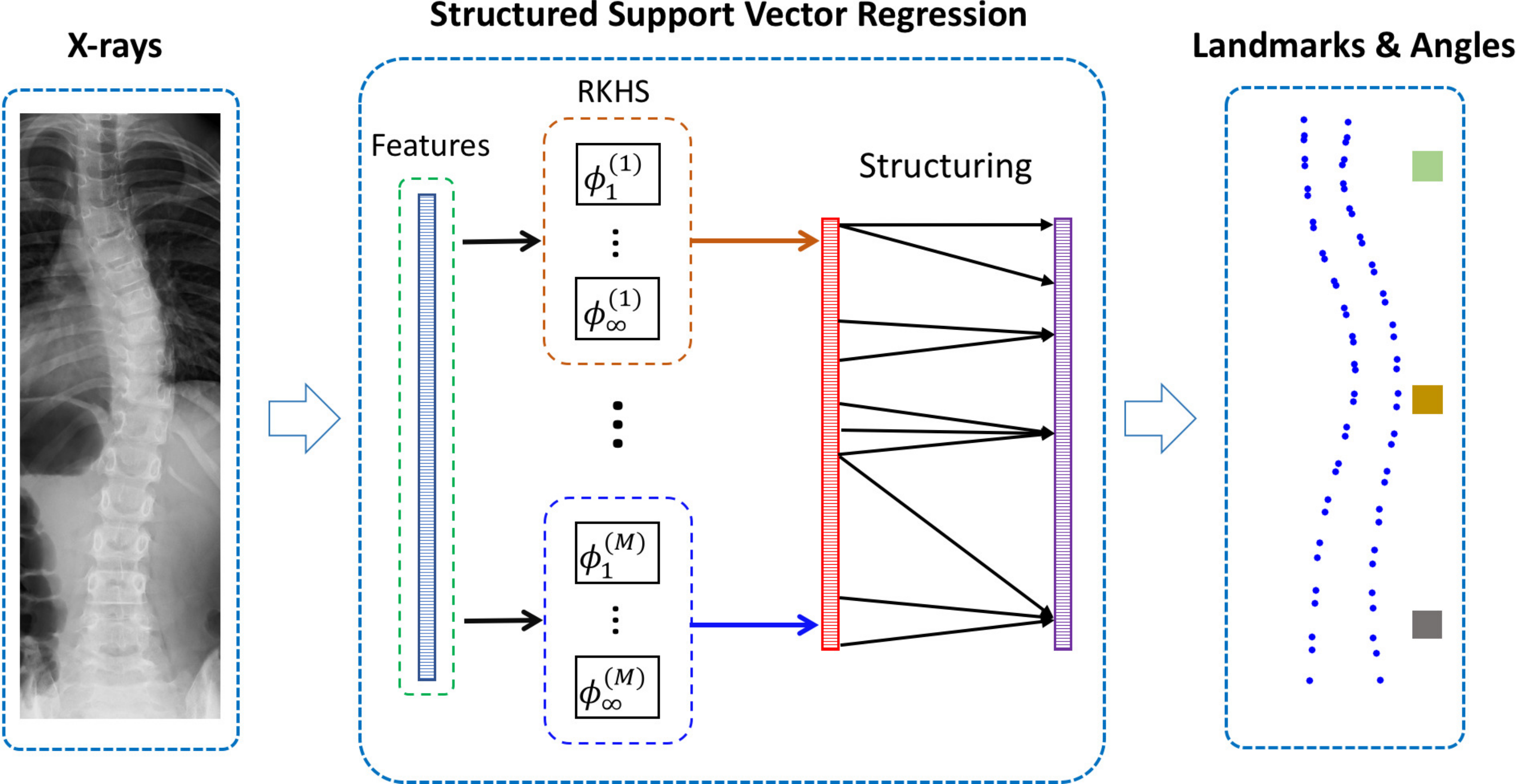}}
\end{center}
\vspace{-8mm}
   \caption{The framework of the structured support vector regression (S$^2$VR).}
\label{fig:framework}
\end{figure}

\vspace{-2mm}
Within the proposed S$^2$VR, we extract image features of a X-ray that denoted as $\mathbf{x}_i \in \mathbb{R}^{\mathrm{d}}$. The coordinates and angles are represented by $\mathbf{y}_i \in \mathbb{R}^{\mathrm{q} }$ and $\mathbf{y}_i = [h_1,...,h_c,v_1,...v_c,a_1,a_2,a_3]$, $c$ is the number of landmarks, $h_i$ and $v_i$ are the horizontal and vertical axises of $i$-th landmark point. In addition, the number of Cobb angles is $3$, thus, $\mathrm{q} = 2c+3$. The regression task is to predict the coordinates and the Cobb angles from the input features. The framework of S$^2$VR is illustrated in Fig~\ref{fig:framework}.

\subsubsection{SVR for Nonlinear Input-output Relationship Learning}

Support vector regression (SVR) has been widely used to handle complex nonlinear relationships. Its effectiveness is guaranteed by the sparseness of solutions. The SVR model takes the following optimization formulation:

\begin{equation}
\argmin L(W,\mathbf{b}) =  \frac{1}{2}|\!|W|\!|^2 +  \tau \sum^N_{i=1}\nu(u_i)
\end{equation}
where \vspace{-3mm}
\begin{equation}
\nu(u_i) = \begin{cases} 0 &u_i < \varepsilon \\
 (u_i - \varepsilon)^2 & u_i \geq \varepsilon \end{cases},
\end{equation}
$u_i  = |\!| \mathbf{y}_i -  f_p(\mathbf{x}_i)|\!|^2$, $f_p(\mathbf{x}_i)= W\phi(\mathbf{x}_i) + \mathbf{b}$ , $|\!|\cdot|\!|$ is $\ell^2$-norm, $\phi : \mathbb{R}^{d} \rightarrow \mathbb{R}^{\mathrm{H}} $ is a nonlinear transformation to a higher $\mathrm{H}$-dimensional space,  $ W \in \mathbb{R}^{q \times \mathrm{H}}$, $\mathbf{b} \in \mathbb{R}^{q}$, $\phi( \mathbf{x}_i) \in \mathbb{R}^{\mathrm{H} }$, $ \mathbf{y}_i \in \mathbb{R}^{q}$, $\tau$ and $N$ denote a positive constant of the tuning parameter and the number of training samples, respectively.

\vspace{-3mm}
\subsubsection{S$^2$VR for Explicit Inter-output Correlation Learning}

We introduce a structure matrix $S, S \in \mathbb{R}^{q \times q}$ to capture the intrinsic correlations of outputs (angles and landmarks). The prediction model becomes $f(\mathbf{x}_i) = S(W\phi( \mathbf{x}_i) + \mathbf{b})$ and the S$^2$VR model can be defined as:

\begin{equation}
L(W,\mathbf{b},S) =  \frac{1}{2}|\!|W|\!|^2 +  \tau \sum^N_{i=1}\nu(u_i) + \lambda |\!|S^\top|\!|_{2,1},
\end{equation}
where $u_i  = |\!| \mathbf{y}_i -  f(\mathbf{x}_i)|\!|$, $|\!|S |\!|_{2,1} = \mathlarger{ââ\sum}_{i=1}^q \sqrt{\sum_{j=1}^q S_{ij}^2} $ and $\lambda$ is the tuning parameter.

\textbf{Advantages}: The appealing property of structure matrix $S$ via $\ell_{2,1}$-norm regularization is to encourage multiple outputs to share the similar sparsity patterns of parameters. Therefore, the structure, such as spatial correlations which share similar specific patterns, can be captured for the robust prediction of angles and coordinates.

\vspace{-3mm}
\subsubsection{Manifold Regularization for S$^2$VR}

To further improve the performance of the proposed algorithm, we consider incorporating a manifold regularization \cite{belkin2006manifold} term based on a graph Laplacian to model the local geometrical structure of the output space. It has been shown that learning performance can be significantly enhanced if the geometrical structure is exploited and the local invariance is considered.

We adopt the loss function with the additional regularization as:
\begin{equation}
\label{ms2vr}
L(W,\mathbf{b},S) = \frac{1}{2}|\!|W|\!|^2 + \frac{\gamma}{2} \tr(FGF^\top) +  \tau \sum^n_{i=1}\nu(u_i) + \lambda |\!|S^\top|\!|_{2,1},
\end{equation}
where $ G \in \mathbb{R}^{N \times N}$ is the graph Laplacian matrix\cite{von2007tutorial} and $F = [f(\mathbf{x}_1),...,f(\mathbf{x}_N)]^\top$.

Here we select the unnormalized graph Laplacian matrix given by $G= \hat{D} - E$ to represent the labels $\mathbf{y}_i|_{i=1:N}$ of data in form of the similarity graph. We simply connect all points with positive similarity with each other, and we weight all edges by $E_{ij}$. The similarity function is the Gaussian function $E_{ij} = \exp(-|\!| \mathbf{y}_i-\mathbf{y}_j|\!|^2/(2\rho^2))$, where the parameter $\rho$ is the width of the function. $\hat{D}$ is the diagonal matrix given by $\hat{D}_{ii} = \sum^N_{j=1}E_{ij}$.

\textbf{Advantages}: The manifold regularization exploits the geometry of the distribution of output data. In particular, due to the high ambiguity and variability of image appearance, distributions of input features and output angles (landmarks) could be inconsistent in estimation. Therefore, by leveraging the strength of manifold regularization, we ensure the smoothness of solution with respect to the output space.

\subsection{Kernelization}
Since the relationship between inputs of image features and outputs of angles (landmarks) is complex and highly nonlinear, we kernelize the (\ref{ms2vr}) to conduct nonlinear regression.   

\vspace{-3mm}
\subsubsection{Kernel S$^2$VR}
Before kernelizaiton, we rewrite (\ref{ms2vr}) in term of traces:
\begin{equation}
\label{trs2vr}
L(W,\mathbf{b},S) = \frac{1}{2}\tr (W^\top W) + \frac{\gamma}{2} \tr(FGF^\top) +  \tau \tr(UE^\top E) + \lambda |\!|S^\top|\!|_{2,1},
\end{equation}
where $E = [\mathbf{e}_1,...,\mathbf{e}_N]$, $\mathbf{e}_i = \mathbf{y}_i - S(W\phi (\mathbf{x}_i))+\mathbf{b})$, and $U$ is the diagonal matrix given by
\begin{equation}
    U_{ii} = \begin{cases} 0 &u_i < \varepsilon \\
    1 & u_i \geq \varepsilon \end{cases}.
\end{equation}

Assume that the mapping function $\phi(\mathbf{x}_i)$ is in some RKHS of infinite dimensionality. According to the Representer Theorem \cite{kimeldorf1970correspondence}, $W = \boldsymbol \beta \boldsymbol \Phi^\top (X)$, where $\boldsymbol \beta \in \mathbb{R}^{q \times N}$ and $X=[\mathbf{x}_1,\dots,\mathbf{x}_N]$. The kernel matrix is defined as $K = \boldsymbol \Phi^\top (X) \boldsymbol \Phi(X)$. We rewrite (\ref{trs2vr}) with the kernel version:
\begin{equation}
     Q(\boldsymbol\beta,S) = \frac{1}{2}\tr(\boldsymbol \beta K {\boldsymbol \beta}^\top ) + \frac{\gamma}{2} \tr(S \boldsymbol \beta KGK {\boldsymbol \beta}^\top  S^\top ) + \frac{1}{2}\tr(UE^\top E) +  \lambda |\!|S^\top|\!|_{2,1},
\label{kers2vr}
\end{equation}
where \vspace{-3mm}
\begin{equation}
   E = Y - S\boldsymbol \beta K.
   \label{slE}
\end{equation}

\vspace{-5mm}
\subsubsection{Learning by Kernel Alignment}
\label{nka}
We focus on learning a cone combination of given base kernels $K_m|_{i=1:M}$ to obtain an efficient kernel. Therefore, for the summation formulation of $K = \sum^M_{m=1} \omega_{m} K_m$, we propose learning the weight coefficients $\boldsymbol\omega$ associated with basic kernels in a supervised way by kernel alignment, which has shown great effectiveness in learning the optimal combination of multiple kernels \cite{cristianini2002kernel,cortes2010two}.

The core idea of kernel alignment is to align an input kernel $K$ to a target kernel $K_T$ by maximizing the similarity or the degree of agreement between them. Specifically, the alignment between kernels is defined as
\begin{equation}
A(K,K_T) = \frac{\langle K, K_T \rangle_F}{\sqrt{\langle K, K_T \rangle_F \langle K_T, K_T \rangle_F}}.
\label{alignment}
\end{equation}
Intuitively, the measurement of alignment can be viewed as the cosine of the angle between two bi-dimensional vectors $K$ and $K_T$.
Kernel target alignment offers a best-suited way to obtain the weight coefficients in $\boldsymbol\omega$. We now introduce the kernel alignment formulation to learn the kernel in S$^2$VR. We would like to maximize the alignment between the target kernel matrix $K_T$ and the kernel $K_{\boldsymbol\omega}$, and based on (\ref{alignment}), we have the following optimization problem
\begin{equation}
\boldsymbol\omega^* = \argmax A(K_{\boldsymbol\omega},K_T) = \argmax \frac{\tr(K_{\boldsymbol\omega} K_T)}{\sqrt{\tr(K_{\boldsymbol\omega} K_{\boldsymbol\omega})}}.
\label{kt}
\end{equation}
The target kernel matrix $K_T$ is constructed by defining the target kernel $K_T = Y^\top Y$, where $Y=[\mathbf{y}_1,\dots,\mathbf{y}_N]$.

As indicated in \cite{cortes2010two}, to obtain the high correlation between performance and kernel alignment, it is necessary to center all kernel matrices $K_m$ before alignment. Let $[K_m]_{ij}$ denote the element in $K_m$ and the centered kernel matrix can be computed by
\begin{equation}
\begin{aligned}
\big[\bar{K}_m \big]_{ij}= &\big[K_m\big]_{ij} -\frac{1}{N} \sum_{i=1}^N \big[K_m\big]_{ij} -\frac{1}{N} \sum_{j=1}^N \big[K_m\big]_{ij} + \frac{1}{N^2} \sum_{i,j=1}^N \big[K_m\big]_{ij}.
\end{aligned}
\end{equation}

We can further equivalently rewrite the objective function in (\ref{kt}) as follows:
\begin{equation}
\label{equation:algmax}
\mathbf{\boldsymbol\omega}^* =  \argmax_{|\!|\mathbf{\boldsymbol\omega}|\!|=1, \mathbf{\boldsymbol\omega} \geq 0} \frac{ \mathbf{\boldsymbol\omega}^\top  \mathbf{\boldsymbol\alpha} \mathbf{\boldsymbol\alpha}^\top  \mathbf{\boldsymbol\omega} }{ \mathbf{\boldsymbol\omega}^\top  V \boldsymbol\omega}
\end{equation}
where $\boldsymbol\omega \geq 0$ guarantees the positive definiteness, $|\!|\mathbf{\boldsymbol\omega}|\!|=1$ is a regularization term, for $i,j \in \{1, \cdots,M \}$, $\boldsymbol\alpha$ is defined by $\boldsymbol\alpha_i = \tr \big(\bar{K}_{i} K_T \big)$ and the matrix $V$ is defined by $V_{ij}=\tr \big(\bar{K}_{i} \bar{K}_{j} \big)$.

This alignment maximum problem in (\ref{equation:algmax}) can be reduced to a simple quadratic programming (QP) problem \cite{wrightnumerical} as shown in the Proposition~\ref{prop1}, which does not require the inversion of $V$ in (\ref{equation:algmax}) and can be solved efficiently.

\begin{prop}
Let $\mathbf{q}^*$ be the solution of the following QP:
\begin{equation}
\label{qp}
\mathbf{q}^* =  \argmin_{\mathbf{q} \geq 0}  \mathbf{q}^\top V \mathbf{q} -2\mathbf{q}^\top \boldsymbol\alpha.
\end{equation}
Then, the solution $\mathbf{\boldsymbol\omega}^*$ of the alignment maximization problem (\ref{equation:algmax}) is given by $
\mathbf{\boldsymbol\omega}^* =\frac{\mathbf{q}^*}{|\!|\mathbf{q}^*|\!|}.$
\label{prop1}
\end{prop}
\begin{proof}
The proof can be referred to the proof of Proposition 3 in \cite{cortes2010two}.
\end{proof}

\textbf{Advantages}: The kernels in S$^2$VR are learned in a supervised way via kernel target alignment, which increases the discriminative ability of the kernels. Also, it is efficient that parameters in kernel functions are learned automatically rather than tuned by cross validation.


\subsection{Alternating Optimization}

S$^2$VR is efficiently optimized by using a new alternating optimization method. The proposed method is fast, whose convergence is theoretically analyzed. Minimization (\ref{kers2vr}) can be decomposed into two sub-problems with respect to the kernel coefficients $\boldsymbol \beta$ and the structure matrix $S$.


\textbf{Fixing $S$}, the minimization problem with respect to $\boldsymbol \beta$ can be optimized by iteratively reweighted least squares (IRWLS) \cite{sanchez2004svm}. We construct a quadratic approximation of (\ref{trs2vr}) about $u_i$:
\begin{equation}
     Q(W,\mathbf{b}) = \frac{1}{2}\tr(W^\top W) + \frac{\gamma}{2} \tr(FGF^\top ) + \frac{1}{2}\tr(DE^\top E) + \tau T,
 \label{QA}
\end{equation}
where $D$ is the diagonal matrix given by
\begin{equation}
    D_{ii} = \frac{\tau}{u_i}\left. \frac{dL(u)}{du} \right|_{u_i}= \begin{cases} 0 &u_i < \varepsilon \\
    \frac{2\tau(u_i-\varepsilon)}{u_i} & u_i \geq \varepsilon \end{cases}
    \label{cpai}
\end{equation}
and $\tau T$ is a sum of constant terms that do not depend either on $W$ or $\mathbf{b}$.

\noindent
We modify (\ref{QA}) in the kernel version:
\begin{equation}
     Q(\boldsymbol\beta) = \frac{1}{2}\tr(\boldsymbol\beta K {\boldsymbol\beta}^\top ) + \frac{\gamma}{2} \tr(S \boldsymbol\beta KLK {\boldsymbol\beta}^\top  S^\top ) + \frac{1}{2}\tr(DE^\top E) + \tau T,
\label{ker:vn}
\end{equation}
where $ E = Y - S\boldsymbol\beta K$.

\noindent
To obtain $ \boldsymbol\beta $, we can equate its gradient to zero:
\begin{align}
     & \frac{\partial Q}{\partial \boldsymbol\beta} = \boldsymbol\beta K - \gamma S^\top S\boldsymbol\beta KGK - S^\top (Y - S \boldsymbol\beta K )DK = 0, \\
     & {(S^\top S)}^{-1} \boldsymbol\beta +  \boldsymbol\beta  K( \gamma G+D) = S^{-1}YD.
     \label{slbeta}
\end{align}
It is the standard Sylvester equation ($A\boldsymbol\beta +\boldsymbol\beta B=C$, where $A = {(S^\top S)}^{-1}$, $B= K( \gamma G+D)$ and $C=S^{-1}YD $).


\noindent
We minimize (\ref{ker:vn}) by constructing a descending direction using the optimal solution of (\ref{slbeta}), and the next step solution is computed by a line search algorithm \cite{wrightnumerical}. Inspired by the IRWLS, we propose the adapted IRWLS procedure summarized in Algorithm~\ref{alg:A}.

\begin{algorithm}[t]
\caption{Adapted IRWLS}
\label{alg:A}
\begin{algorithmic}[1]
\STATE {Initialization: set $k=0$, compute $D^k$ using (\ref{cpai})}
\REPEAT
\STATE Compute $ \boldsymbol\beta^k$ in (\ref{slbeta}) by using
the Sylvester equation, and denote it as $\boldsymbol\beta^\ell$. Get the descending direction as $\Delta^k=\boldsymbol\beta^\ell - \boldsymbol\beta^k$.

\STATE Update $\boldsymbol\beta^{k+1} = \boldsymbol\beta^k + \eta^k\Delta^k$, and the step size $\eta^k$ is computed by using the backtracking algorithm.
\STATE Obtain $D^{k+1}$.
\STATE Set $k=k+1$.
\UNTIL{Convergence}
\end{algorithmic}
\end{algorithm}

\begin{algorithm}[t]
\caption{Iteratively Computing $S$}
\label{alg:B}
\begin{algorithmic}[1]
\STATE {Initialization: set $k=0$}
\REPEAT
\STATE Compute the diagonal matrix $P^k$.
\STATE Update $S^{k+1}$ using (\ref{sls}).
\STATE Set $k=k+1$.
\UNTIL{Convergence}
\end{algorithmic}
\end{algorithm}

\textbf{Given fixed $\boldsymbol\beta$}, the derivative of (\ref{kers2vr}) respect to $S$ is computed as:
\begin{align}
     & \frac{\partial Q}{\partial S} =  2\lambda SP - \tau(Y-S \boldsymbol \beta K)D(\boldsymbol \beta K)^\top  + \gamma S \boldsymbol\beta KGK {\boldsymbol\beta}^\top  = 0 \\
     & S = \tau YDK{\boldsymbol\beta}^\top  {\left(2\lambda P + \tau \boldsymbol\beta KDK{\boldsymbol\beta}^\top  + \gamma \boldsymbol\beta KGK {\boldsymbol\beta}^\top \right)}^{-1},
     \label{sls}
\end{align}
where $S$ is a diagonal matrix with $ P_{jj} = \frac{1}{2||S_j||_2}, P \in \mathbb{R}^{q \times q} $ and $ S_j $ is the column vector of $ S $.

\noindent
We compute the derivative of $S$ using (\ref{sls}). The calculation of $P$ is based on $S$. Therefore, the new $S$ is computed by using the current $S$. The procedure is summarized in Algorithm \ref{alg:B}. Therefore, minimization (\ref{kers2vr}) can be solved by alternatively   optimizing $\boldsymbol\beta$ and $S$.

\textbf{Proof of Convergence}: The fast convergence for optimization ensures the efficiency of our method. We provide the theoretical analysis with rigorous proof for the convergence of the optimization algorithm. Since the optimization is conducted in an alternating iterative way, we first prove the convergence for the adapted IRWLS procedure for optimizing $\boldsymbol\beta$.

\begin{thm}
$Q(\boldsymbol\beta)$ in (\ref{ker:vn}) bounds together all outputs and monotonically decreases in each iteration of Algorithm~\ref{alg:A}ã
\label{thm:1}
\end{thm}
\begin{proof}
The proof can be referred to the proof of Appendix in \cite{sanchez2004svm}.
\end{proof}

Now that the convergence for the adapted IRWLS procedure is guaranteed in Theorem~\ref{thm:1}, the following theorem shows the convergence of the alternating iterative method.

\begin{thm}
$Q(\boldsymbol\beta,S)$ in (\ref{kers2vr}) is bounded from below and monotonically decreases with each alternating iterative optimization step for $\boldsymbol\beta$ and $S$.
\label{thm:2}
\end{thm}
\begin{proof}
Due to the sum formation of $Q(\boldsymbol\beta,S)$ in (\ref{ker:vn}), we have the bound of $Q(\boldsymbol\beta,S) \geq 0$. Let $\boldsymbol\beta^{(t)}$ and $S^{(t)}$ denote $\boldsymbol\beta$ and $S$ in the $t$-th iteration respectively. Then, $\boldsymbol\beta^{(t)}$ and $S^{(t)}$ are computed by $\boldsymbol\beta^{(t)} \leftarrow \argmin_{\boldsymbol\beta} Q(\boldsymbol\beta,S^{(t-1)})$ and $S^{(t)} \leftarrow \argmin_S Q(\boldsymbol\beta^{(t-1)},S)$. Since $Q(\boldsymbol\beta,S^{(t-1)}) \geq Q(\boldsymbol\beta,S^{(t)})$ has been proved in \cite{nie2010efficient}, we obtain the following inequality:
\begin{equation}
     \cdots \geq Q(\boldsymbol\beta^{(t-1)},S^{(t-1)}) \geq Q(\boldsymbol\beta^{(t)},S^{(t-1)}) \geq Q(\boldsymbol\beta^{(t)},S^{(t)}) \geq \cdots .
\label{inequation}
\end{equation}
$Q(\boldsymbol\beta^{(t)},S^{(t)})$ is monotonically decreasing as $t \rightarrow +\infty$. Therefore, the convergence of the optimization of $Q(\boldsymbol\beta,S)$ is proved.
\end{proof}

\subsection{Prediction}
In the testing stage, given a new input $\mathbf{x}^{(t)}$, the predicted angles and landmarks are given by $\hat{\mathbf{y}}^{(t)}=S\boldsymbol\beta K^{(t)}$, where $K^{(t)} = \sum_{m=1}^M \omega_{m} K_m^{(t)}$, $K_m^{(t)} = \boldsymbol \Phi^\top (\hat{X}) \phi(\mathbf{x}^{(t)})$ and $\hat{X}$ is the matrix composed of the support vectors.


\section{Experiments}
S$^2$VR has been validated on the spinal X-ray dataset with a large number of subjects. Extensive experiments show that our method with significant effectiveness consistently outperforms the baseline methods, which can be practically used in clinical scoliosis analysis.

\vspace{-3mm}
\subsection{Datasets and Implementation Details}
\vspace{-2mm}

The posteroanterior spinal X-ray dataset is composed of 439 samples from different individuals. Since the cervical vertebrae (the vertebrae of the neck) are seldom involved in spinal deformity \cite{brien2008manual}, we select 17 vertebrae composed of the thoracic spine and lumbar spine for spinal shape characterization. And each vertebra is located by four landmarks with respect to four corners. The landmarks of a spine consist of $c=68$ points which are annotated manually, while the Cobb angles with $3$ values (from top to bottom: TA, MA, BA) are measured by human experts. We choose histogram of oriented gradient (HOG) descriptor \cite{dalal2005histograms} as the input of the model due to its strength in local representation. The  Gaussian Kernel is selected as the basic kernel and the $\sigma$ parameter is in the range of $[0.1, 1]$. Our model has been validated using the leave-one-out cross validation scheme.

\subsection{Results}
\begin{figure}
 \begin{center}
        \setbox0\hbox{\includegraphics[width=1.379cm,height=3.5cm]{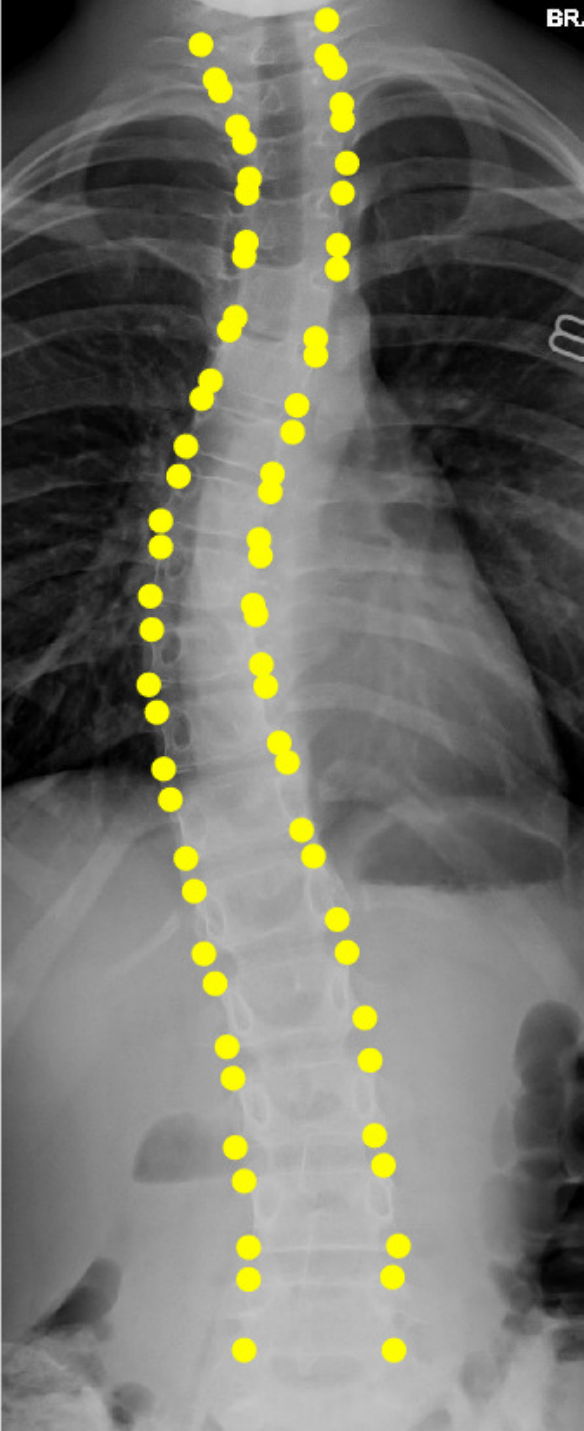}}
        \setbox2\hbox{\includegraphics[width=1.379cm,height=3.5cm]{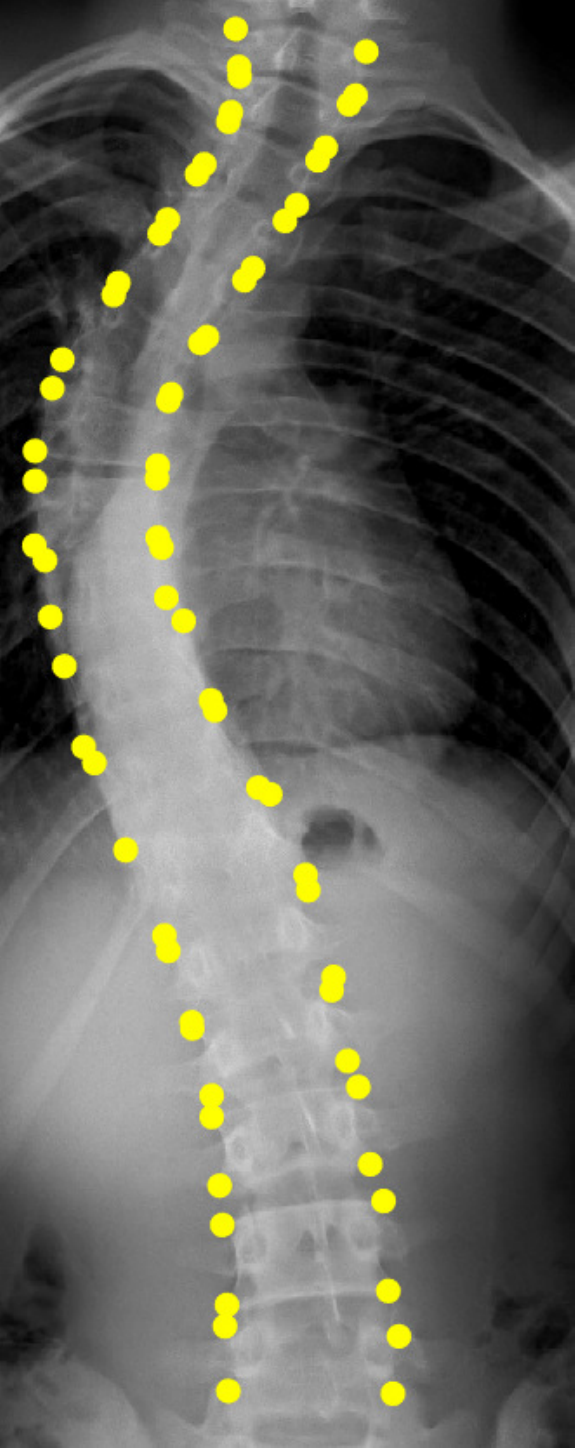}}
        \setbox4\hbox{\includegraphics[width=1.379cm,height=3.5cm]{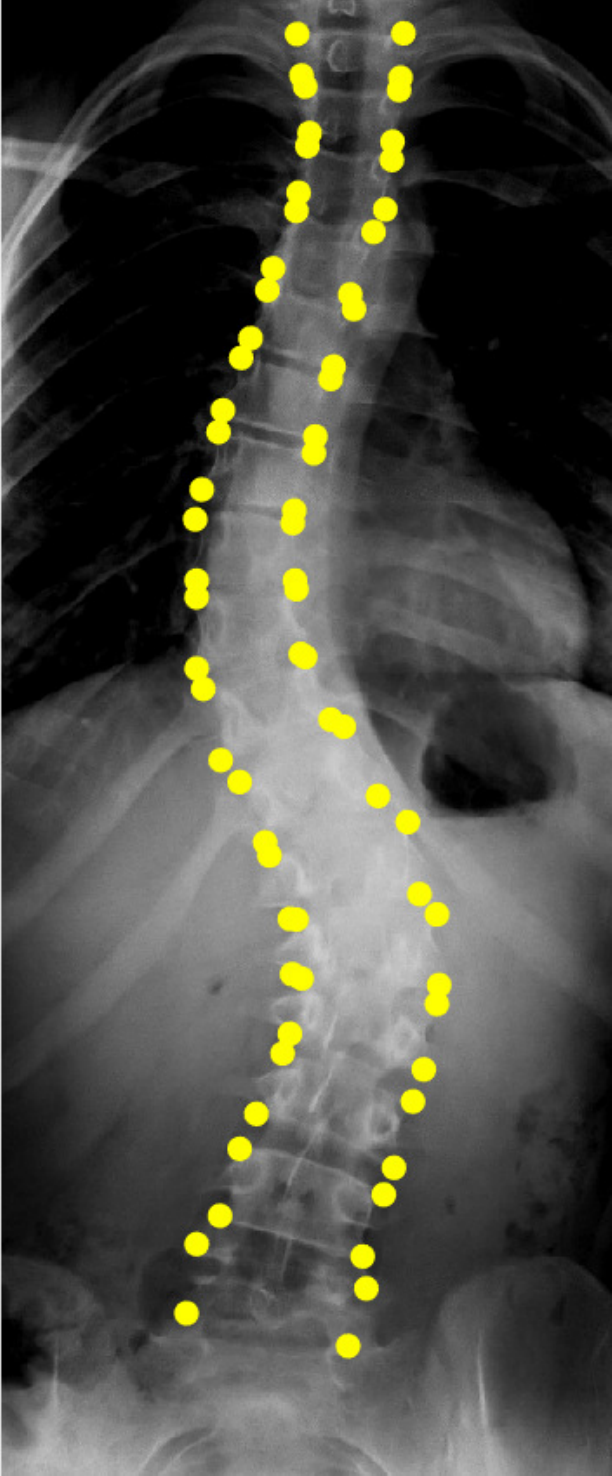}}
        \setbox6\hbox{\includegraphics[width=1.379cm,height=3.5cm]{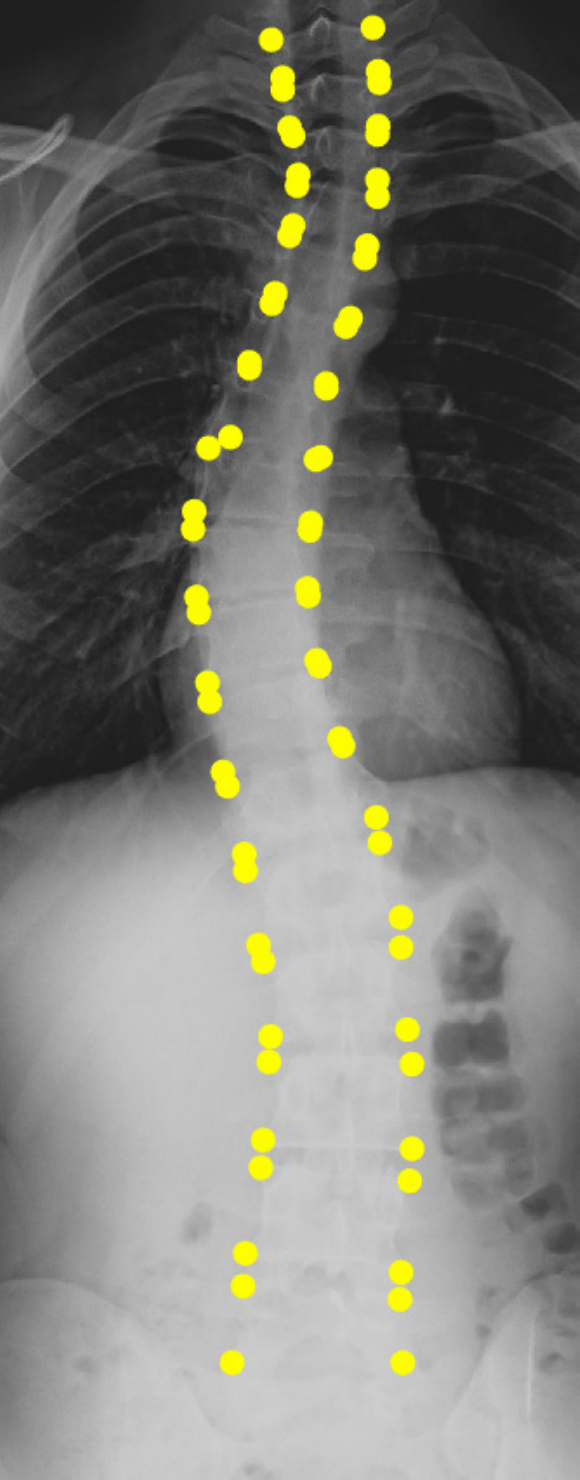}}
        \setbox8\hbox{\includegraphics[width=1.379cm,height=3.5cm]{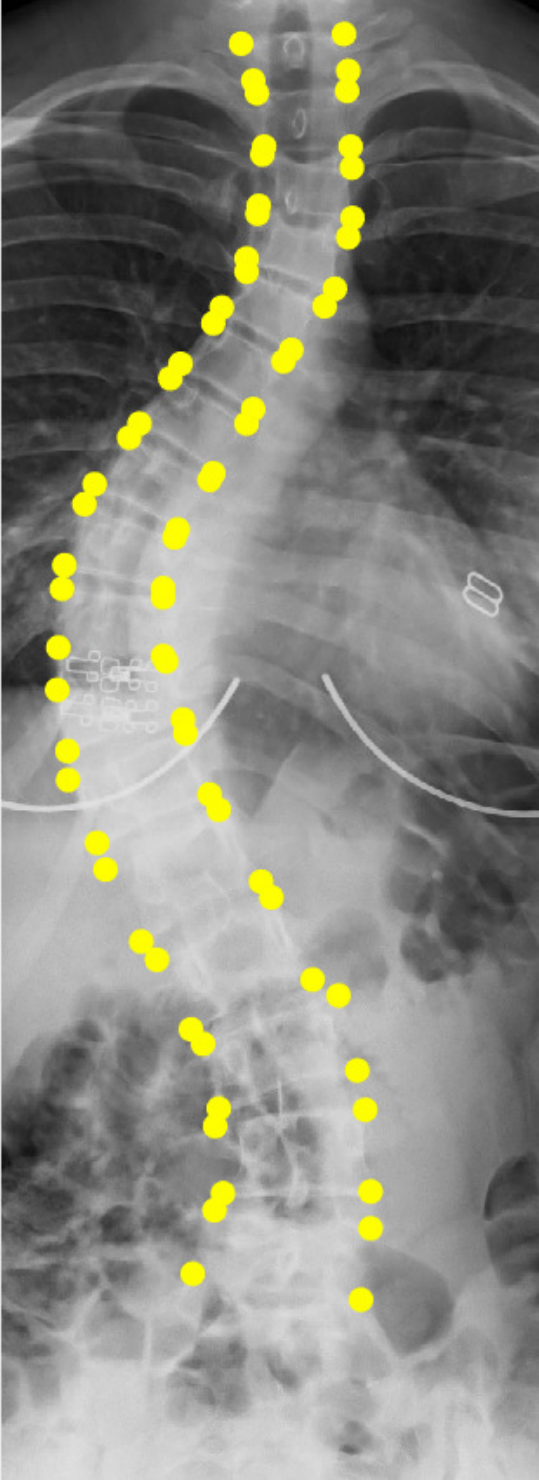}}

        \noindent
        \parbox{.15\textwidth}{%
                \centering
                \unhbox0
        }%
        \hfil
        \parbox{.15\textwidth}{%
                \centering
                \unhbox2
        }%
        \hfil
        \parbox{.15\textwidth}{%
                \centering
                \unhbox4
        }%
        \hfil
        \parbox{.15\textwidth}{%
                \centering
                \unhbox6
        }%
        \hfil
        \parbox{.15\textwidth}{%
                \centering
                \unhbox8
        }

\caption{Our method overcomes huge variations and high ambiguities and achieves high accuracy in landmark detection.}
\label{fg:ldm}
\end{center}
\vspace{-8mm}
\end{figure}

The S$^2$VR achieves desirable performance on the spinal X-ray dataset and demonstrates great effectiveness for the Cobb angle estimation and landmark detection. We show the qualitative results of landmark detection in Fig~\ref{fg:ldm}. We also compare with two baseline methods, which shows the strength of our method on the spinal curvature analysis.

\textbf{Comparison}. As demonstrated in Tables~1 and 2, we compare with two baseline methods, i.e. support vector regression (SVR) \cite{sanchez2004svm}, shape regression machine (SRM) \cite{zhou2010shape} about the relative root mean squared error (RRMSE) and the correlation coefficient, which shows the effectiveness of our method. For vectors of the length of $l$, let $\hat{y}$, $y$, $\bar{Y}$ denotes the ground truth, the estimated vector, the mean of target variable $Y$ over train data, respectively, the RRMSE is defined as
\begin{equation}
R_r = \frac{\sum_{i=1}^{l}\hat{y}_i-y_i}{\sum_{i=1}^{l}\bar{Y}_i-y_i}.
\end{equation}

The S$^2$VR achieves the lowest average RRMSE of $21.63$ and the highest correlation coefficient of $92.76$. Since there are strong correlations between outputs, explicit correlation learning in the S$^2$VR can faithfully exploit the structure information for a better result. Also, the learned and more robust kernel is powerful enough to handle the correspondence between the image appearance and high-level semantic concepts. We also compare the separated prediction with joint prediction. As shown in two tables, the joint prediction consistently outperforms than the separated one.

\begin{table*}
\label{etb1}
\vspace{-7mm}
\begin{center}
\caption{The comparison of the average RRMSE ($\%$).}
\begin{tabular}{p{3cm}  p{3cm}    c}
\hline
\footnotesize{Method}     & Angles    &Angles \& Landmarks\\
\hline
SVR \cite{sanchez2004svm}  &$25.01$&   $23.72$\\
SRM \cite{zhou2010shape}   &$24.83$&   $23.35$\\
\textbf{S$^2$VR} (Ours)    &$23.69$&   $21.63$\\
\hline
\end{tabular}
\end{center}
\vspace{-4mm}
\end{table*}

\begin{table*}
\label{etb2}
\vspace{-4mm}
\begin{center}
\caption{The comparison of the correlation coefficient ($\%$) of three angles.}
\begin{tabular}{p{3cm}  p{3cm}    c}
\hline
\footnotesize{Method}    &Angles    &Angles \& Landmarks\\
\hline
SVR \cite{sanchez2004svm}  &$90.83$&   $91.95$\\
SRM \cite{zhou2010shape}   &$91.30$&   $92.13$\\
\textbf{S$^2$VR} (Ours)    &$91.94$&   $92.76$\\
\hline
\end{tabular}
\end{center}
\vspace{-7mm}
\end{table*}

\begin{figure}
\centerline{\includegraphics[width=1.15\linewidth]{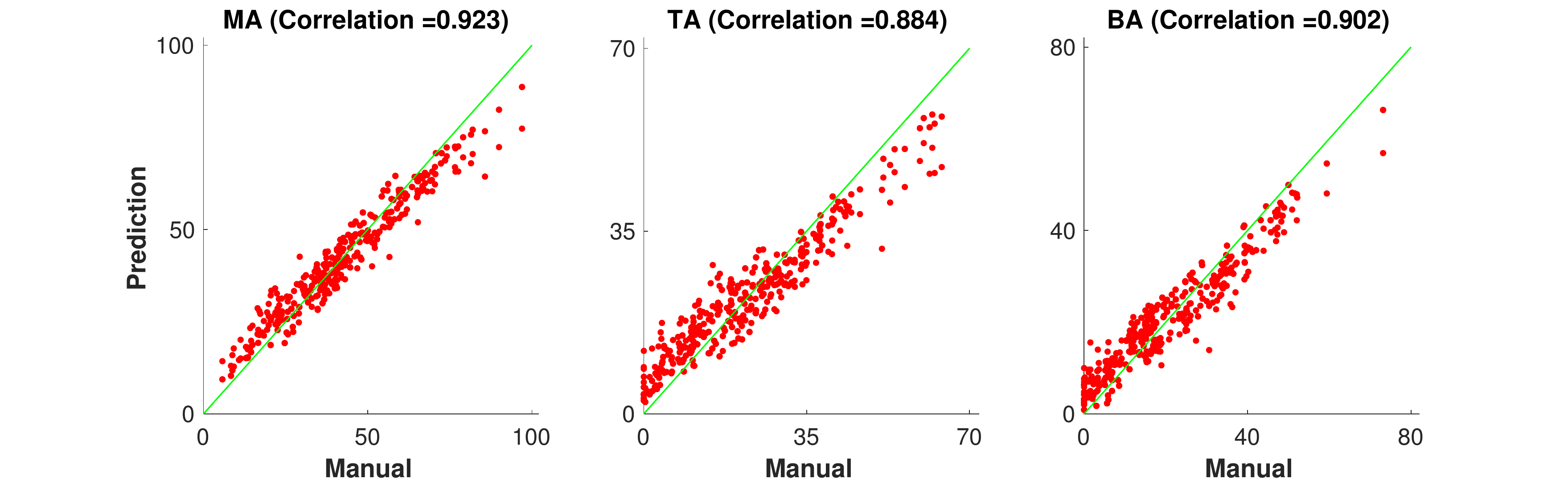}}
\caption{The correlation coefficients between three angles predicted by the proposed method and ground truth for the MA, TA and BA, respectively. The method achieves a high correlation coefficient with manual measurement.}
\label{fg:1}
\vspace{-7mm}
\end{figure}

\textbf{Effectiveness}. The effectiveness of the proposed method is demonstrated by the outstanding performance on the spinal X-ray dataset. The estimated Cobb angles by the proposed method are compared with those by manual measurement. The correlations between estimated angles and ground truth are depicted in Fig~\ref{fg:1}, respectively. The proposed method achieves a correlation coefficient of $0.923$ for the middle angle and can yield $0.884$ and $0.902$ for the rest two angles.

\vspace{-3mm}
\section{Conclusion}
\vspace{-3mm}
In this paper, we propose structured support vector regression (S$^2$VR) model to directly predict the Cobb angles and landmarks in spinal X-rays from image features. The proposed S$^2$VR consists of non-linear mapping and explicit structure modeling, which can handle the highly nonlinear relationship between image features and quantitative evaluation parameters and explicitly learn the inter-output corrections. Moreover, the manifold regularization is introduced for the smooth solution. To obtain a discriminative kernel, we propose to learn the kernel in S$^2$VR by kernel target alignment, which can leverage the strength of supervised kernel learning. Extensive experiments on spinal X-ray dataset show the great effectiveness of our method compared with two baseline methods.

\textbf{Acknowledgments.} The work is supported by NSFC Joint Fund with Guangdong under Key Project No. U1201258, National Natural Science Foundation of China under Grant No. 61573219, and the Fostering Project of Dominant Discipline and Talent Team of Shandong Province Higher Education Institutions.

{\small
\bibliographystyle{elsarticle-num}
\bibliography{ipmi}
}

\end{document}